\pdfoutput=1
\documentclass[11pt]{article}

\usepackage[]{ACL2023}

\usepackage{tipa}
\usepackage{times}
\usepackage{latexsym}
\usepackage{multirow}   %  table
\usepackage{graphicx}   %  .png
\usepackage{enumitem}
\usepackage{amsmath}    %  symbols
\usepackage{amssymb}    %  symbols
\usepackage{latexsym}
\usepackage[inkscapelatex=false]{svg}
\usepackage{url}    % \url
\usepackage{colortbl, booktabs} % To thicken table lines
\usepackage{makecell} % automatic words wrap
\usepackage{color}  %  color
\usepackage{array}
\usepackage{booktabs}
\usepackage{float}
\usepackage{hyperref}   %  \href{url}{caption}
\usepackage[all]{nowidow}   % prevents the appearance of an ugly single line on the second page.
\usepackage[subtle]{savetrees} % The goal of the savetrees package is to pack as much text as possible onto each page

\usepackage{algorithm}
\usepackage{algorithmic}

\usepackage[T1]{fontenc}
\usepackage{microtype}

\usepackage{inconsolata}

\usepackage[T1]{fontenc}

\title{TwistList: Resources and Baselines for Tongue Twister Generation}
\author{Tyler Loakman\textsuperscript{1}\footnotemark[1], Chen Tang\textsuperscript{2}\footnotemark[1]  ~and Chenghua Lin\textsuperscript{1}\footnotemark[2]\\
  \textsuperscript{1}Department of Computer Science, The University of Sheffield, UK \\
  \textsuperscript{2}Department of Computer Science, The University of Surrey, UK \\
  \texttt{\{tcloakman1,c.lin\}@sheffield.ac.uk} \\
    \texttt{chen.tang@surrey.ac.uk} }
  
\begin{document}
\maketitle

%  for authors' footnotes
\renewcommand{\thefootnote}{\fnsymbol{footnote}} 
\footnotetext[1]{Equal contribution.} 
\footnotetext[2]{Corresponding author.} 
\renewcommand*{\thefootnote}{\arabic{footnote}}

\begin{abstract}
Previous work in phonetically-grounded language generation has mainly focused on domains such as lyrics and poetry. In this paper, we present work on the generation of tongue twisters -  a form of language that is required to be phonetically conditioned to maximise sound overlap, whilst maintaining semantic consistency with an input topic, and still being grammatically correct. We present \textbf{TwistList}, a large annotated dataset of tongue twisters, consisting of 2.1K+ human-authored examples. We additionally present several benchmark systems (referred to as \textbf{TwisterMisters}) for the  proposed task of tongue twister generation, including models that both do and do not require training on in-domain data. We present the results of automatic and human evaluation to demonstrate the performance of existing mainstream pre-trained models in this task with limited (or no) task specific training and data, and no explicit phonetic knowledge. We find that the task of tongue twister generation is challenging for models under these conditions, yet some models are still capable of generating acceptable examples of this language type.

\end{abstract}

% =============================== Section 1 ==================================
\section{Introduction}
Phonetically constrained language generation is a primary subarea of computational creativity in natural language generation (NLG), primarily encompassing lyric and poetry generation \cite{tian-peng-2022-zero, wockener-etal-2021-end, xue-etal-2021-deeprapper, zhang-etal-2020-youling, agarwal-kann-2020-acrostic}, as well as pun generation \cite{Sun2022, he-etal-2019-pun, yu-etal-2018-neural}, and continues to prove challenging for myriad reasons. Primarily, such works require the inclusion of phonetic factors such as metre and rhyme, which involves careful consideration of candidate vocabulary on the syllable level, leading to a reduced pool of allowable vocabulary once these constraints are in place.

In this paper, we present work on the generation of \textit{tongue twisters}, a type of phonetically constrained language that is rarely explored in the NLG community.  
%we present work on the generation of a newly studied type of phonetically constrained language, tongue twisters,  
As a form of creative generation, tongue twisters can facilitate numerous useful applications, including: (1) being used as a pedagogical tool \cite{Sugiharto-Japanese, SOMOFF-UNMASTERY, WILSHIRE-1999}; (2) as a source of humorous entertainment stemming from unintentional mispronunciations; (3) as a stylistic device for engaging children in reading (e.g. Dr. Seuss stories \cite{seuss}); (4) as a method of designing memorable slogans and tag lines \cite{guerini-etal-2015-echoes}; and (5) as stimuli in neuroscience/physiology research \cite{wong-broca, oHalloran-apnoea, Kember-dysarthria}. 

% ----------- fig:overview -----------
\begin{figure}[t]
\centering
\includegraphics[width=1\linewidth]{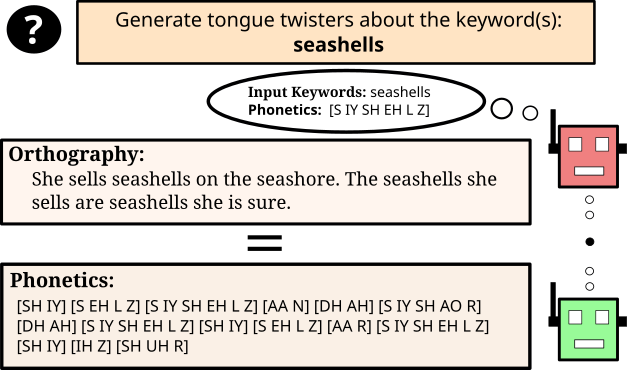}
% \includesvg[width=1.0\columnwidth]{figs/intro.svg}
\caption{Tongue Twister Generation aims to generate an utterance with high levels of phonetic overlap, requiring understanding of semantics, grammar, and phonetics.} 
\label{fig:intro}
\end{figure}
% ----------- end of fig -----------

Tongue twister generation posits unique challenges compared to other generation tasks. One of the most pertinent features of tongue twisters is the presence of high levels of phonetic overlap across tokens \cite{WILSHIRE-1999}. Consequently, whilst other types of creative generation may require only \textit{some} output tokens to consider phonetics (such as rhyme or syllable counts), tongue twisters present an extreme version of this problem where the phonetics of almost all generated tokens must be considered. This leads to a very small vocabulary from which to choose semantically relevant words, and presents further challenges with maintaining grammatical validity.

The only work that we are aware of on tongue twister generation at the time of conducting this research is by
%Recently, generation of such language has been explored by
\citet{pancetta}, who present models that train on  graphemes and phonemes, and take either a starting prompt to be continued, or keywords around which to theme an output. They release \textit{TT-Corp}, a dataset of 644 tongue twisters with parallel non-twister equivalents. We differentiate our work through the release of a dataset that is over 3x larger and which has undergone substantial human quality control. Furthermore, we assess the results of a wider range of popular pre-trained models on this task, including ChatGPT, without explicit injection of phonetic knowledge due to the difficulty in encoding phonetics and the expertise required to utilise phonetic characteristics appropriately. Our experimental results show that most popular pretrained language models (PLMs) rely on pure word repetition to generate tongue twisters, whilst some (i.e. BART) are able to generate more sophisticated examples. Additionally, very large zero-shot models (i.e. ChatGPT) are able to generate convincing tongue twisters almost on-par with human equivalents.

To summarise our contributions, we present:
\begin{itemize}
[noitemsep,topsep=0pt,parsep=0pt,partopsep=0pt,leftmargin=*]
\item \textbf{TwistList}, a large annotated dataset of human-authored tongue twisters, containing 2.1K+ examples with human evaluation of their quality.
\item \textbf{TwisterMisters}, a series of baseline models for tongue twister generation using the most popular state-of-the-art PLMs.
\item Extensive automatic and human evaluation to assess the ability of PLMs to implicitly model the complex phonetic phenomena in tongue twisters.
\end{itemize}
Our code and resources can be accessed at \url{https://github.com/tangg555/TwistList}

\section{Related Works} Previous work in phonetically constrained generation has taken one of two approaches: 1) train a generation model on a collection of in-domain texts, or 2) train a generation model on prosaic out-of-domain text, with constraints imposed at decoding time. For example, \citet{lau-etal-2018-deep} collect 3,355 sonnets to produce novel poetry and train models to generate text in iambic pentameter, whilst \citet{xue-etal-2021-deeprapper} train a rap generation model on 272,839 in-domain examples, infusing knowledge of rhythm afterwards. On the other hand, \citet{van-de-cruys-2020-automatic} train on a subset of CommonCrawl, imposing constraints on topic and rhyme as \textit{a priori} distributions, whilst \citet{tian-peng-2022-zero} train a title-to-keyword module on narrative texts in addition to a sonnet generation model trained on news articles and short stories from Reddit. They imposed literary techniques (simile/metaphor) and metre/rhyme constraints at decoding time, owing to the lack of sufficient training data.\footnote{Additionally, there is often a reluctance in computational creativity to train on examples, under the assumption that the newly generated content will be overly derivative.}

\section{Tongue Twister Generation}

\subsection{Task Definition}

We formulate the task of tongue twister generation as follows: for a given set of keywords, we aim to generate a tongue twister $T$, whereby $T$ comprises a sequence of words $ \{w_1, w_2,... w_{n}\}$. The generated output must satisfy the following constraints: (1) the output should be semantically related to the input keywords; (2) the output should show maximal levels of phonetic overlap across tokens; and (3) the output should be grammatically valid \cite{WILSHIRE-1999}. Of these requirements, phonetic overlap is the most central to defining text as a ``tongue twister''.

% ----------- table dataset info ---------------
\begin{table}[tb]
\scriptsize
\centering
\resizebox{0.99\linewidth}{!}{
\begin{tabular}{l|cccc}
\toprule
\textbf{Dataset}&\textbf{Train}&\textbf{Val}&\textbf{Test}&\textbf{Total}\\
\midrule
\textbf{\# Tongue Twisters}& 1912 & 106 & 107 & 2128 \\
\textbf{Vocabulary Size} & 9556 & 946 & 880 & 10358 \\
\midrule
\textbf{\# Total Phonemes} & 55 & 43 & 46 & 56 \\
\textbf{\# RAKE Keywords} & 3333 & 316 & 288 & 3567\\
\textbf{\# BERTopic Keywords} & 250 & 132 & 160 & 250 \\
\midrule
\textbf{Avg. \# Input Keywords (RAKE)} & 3.16 & 3.32 & 3.01 & 3.16 \\
\textbf{Avg. \# Input Phonemes} & 5.57 & 5.83 & 5.16 & 5.56 \\
\textbf{Avg. Tongue Twister Length (Words)} & 15.01 & 16.59 & 13.54 & 15.01\\
\textbf{Avg. \# Input Phonemes} & 26.06 & 28.25 & 23.50 & 26.04 \\
\bottomrule
\end{tabular}
}
\caption{The Statistics of \textbf{TwistList}.}
\label{tab:data_stat}
\end{table}
% ----------- end of tab ---------------------

% ######################### Section 3.2 ########################
\subsection{TwistList Dataset}
\label{section:construction}
\paragraph{Dataset Construction.}
We present \textbf{TwistList}, an annotated dataset of 2.1K+ human-authored tongue twisters for use by the community. The examples contained therein come from a variety of sources available on the web.\footnote{The source of each tongue twister is stated for each entry.} For each tongue twister, phonetic transcription is provided using the \textit{g2p-en} package,\footnote{\url{https://pypi.org/project/g2p-en/}} in addition to keywords extracted with RAKE and BERTopic to represent the topic of the tongue twister. Following experimentation with both RAKE and BERTopic, only RAKE keywords are used in training due to human preference and issues regarding the use of BERTopic on short texts (where frequently no keywords are extracted).  The main statistics of the dataset are presented in \autoref{tab:data_stat}.

% ----------- tab: simple case study -----------
% VERSION WITH INPUT-OUTPUT PAIRS
\begin{table}[ht]
\centering
\resizebox{\linewidth}{!}{
\begin{tabular}{p{0.2\linewidth}|p{0.80\linewidth}}
\toprule%[2pt]
\textbf{RAKE:} & sells thick socks  \\
%\textbf{Phonemes:} & [S EH1 L Z] [TH IH1 K] [S AA1 K S] \\
\midrule
\textbf{BERTopic:} & short shorts socks sock  \\
%\textbf{Phonemes:} & [SH AO1 R T] [SH AO1 R T S] [S AA1 K S] [S AA1 K] \\
\midrule
\textbf{Twister:} & Seth at Sainsbury's sells thick socks. \\
\textbf{Phonetics:} & [S EH1 TH] [AE1 T] [S EY1 N S B ER0 IY0 Z] [S EH1 L Z] [TH IH1 K] [S AA1 K S]\\
\bottomrule
\end{tabular}
}
\caption{Example from TwistList}
\label{tab:examples}
\end{table}

% ----------- tab: automation metrics -----------
\begin{table*}[t]
\centering \small
\resizebox{0.95\linewidth}{!}{
\begin{tabular}{l|c|cc|ccc|cc|ccc}
\toprule
\textbf{Model} & \textbf{PPL$\downarrow$} & \textbf{B-1$\uparrow$} & \textbf{B-2$\uparrow$}   & \textbf{R-1$\uparrow$} & \textbf{R-2$\uparrow$} & \textbf{R-L$\uparrow$}  & \textbf{PO$\downarrow$} & \textbf{Init-PO$\downarrow$} & \textbf{BS-P$\uparrow$} & \textbf{BS-R$\uparrow$} & \textbf{BS-F$\uparrow$} \\
\midrule
\textbf{GPT-2} & 8.40 & 0.007 & 0.003 & 1.301 & 0.123 & 1.315 & 0.022 & 0.020 & 0.690 & 0.810 & 0.744 \\
\textbf{DialoGPT} & 3.83 & 0.038 & 0.025 & 7.724 & 3.610 & 7.640 & 0.069 & 0.089 & 0.754 & 0.831 & 0.790 \\
\textbf{T5} & 10.16 & 0.057 & 0.038 & 9.701 & 4.573 & 9.574 & 0.689 & 0.727 & 0.795 & 0.818 & 0.806\\
\textbf{BART} & 1.65 & 0.073 & 0.051 & 11.883 & 6.109 & 10.353 & 0.075 & 0.120 & 0.795 & 0.845 & 0.819 \\ 
\midrule
\textbf{ChatGPT} & N/A & 0.200 & 0.137& 36.765 & 20.659 & 33.437 & 0.093 & 0.157 & 0.888 & 0.894 & 0.883 \\ 
\bottomrule
\end{tabular}
}
\caption{\label{tab:auto-evaluation}
Results of Automatic Evaluation. Golden PO = 0.385 and Golden Init-PO = 0.417. Due to the one-to-many issue in creative language generation, we acknowledge that the referenced metrics are imperfect.
}
\end{table*}

\paragraph{Quality Control.}
Quality control on our dataset was performed in multiple ways. Firstly, it was ensured that only sufficiently unique tongue twisters were kept in the dataset, as determined by removing examples with over 90\% word overlap (rather than keeping variants of the same tongue twister, such as ``Peter Piper picked a pickled pepper'' versus ``Peter \textit{the} Piper picked...''). Additionally, non-standard spellings were manually converted to standard US English\footnote{For example, where phonetic spellings or letter substitutions such as ``k'' for ``c'' were used for literary and visual effect, such as ``kwik`` for ``quick``.} to avoid G2P (Grapheme-to-Phoneme conversion) issues.\footnote{\textit{g2p-en} uses the CMU Pronouncing Dictionary to retrieve transcriptions, which is an American English resource.} Similarly, tongue-twisters containing obscure vocabulary (such as medicine and dinosaur names) were excluded to further minimise errors. An annotation platform was developed (see Appendix \ref{apx:QC}), with which 3 human evaluators, who are native speakers of English, were provided with 100 sampled instances from the dataset to rate the quality of the resulting tongue twisters and the associated extracted keywords. The full dataset contains 2,500+ tongue twisters, of which 2,128 are kept for training/development/testing after filtering examples with insufficient extracted keywords and excessive similarity to existing entries.

% ----------- tab: human evaluation-----------
\begin{table}
\centering
\resizebox{0.95\linewidth}{!}{
\begin{tabular}{r|lcc|c}
\toprule[1pt]
\multirow{2}{*}{\textbf{Choices (\%)}} & \multicolumn{4}{c}{\textbf{Sample Quality}}  \\
\cline{2-5} 
& \textbf{High.} & \textbf{Suitable.} & \textbf{Bad.} & \textbf{Kappa}   \\
\midrule
\textbf{RAKE keywords}  & \textbf{82.0} & 18.0 & 0.0 & 0.321 \\
\textbf{BERTopic keywords}  & 15.0 & \textbf{85.0} & 0.0 & 0.445  \\
\textbf{Tongue Twisters}  & \textbf{88.0} & 6.0 & 4.0 & 0.321 \\
\bottomrule[1pt]
\end{tabular}
}
\caption{Kappa refers to Fleiss' Kappa \cite{fleiss1971measuring}. All results achieve fair or moderate agreement. Good tongue twisters that are deemed a bit longer (3\%) or shorter (3\%) than expected are considered "suitable".}
\label{tab:human_eval_dataset}
\end{table}
% ----------- end of tab-----------

To summarise, 3 annotators evaluated the quality of the dataset, where 88\% of assessed tongue twisters were considered high quality, and 6\% considered ``suitable'' (Kappa = 0.321). An example from \textbf{TwistList} is provided in \autoref{tab:examples}. As \autoref{tab:human_eval_dataset} shows, the final dataset can be considered high quality, owing to fair/moderate levels of approval and agreement across evaluators. Demographic information of the evaluators can be found in Appendix \ref{apx:Human Participants}. 
%\textcolor{red}{All remaining tongue twisters and transcriptions were finally reviewed by a trained linguist with expertise in phonetics, who was not involved in the aforementioned evaluation procedure.}

\subsection{Baseline Models}
We present the following baseline models (dubbed \textbf{TwisterMisters}) for the task of tongue twister generation on our TwistList dataset:
\paragraph{Finetuned Baselines.}
For the finetuned baselines, we chose popular models for language generation, including \textbf{GPT-2} \cite{GPT-2}, \textbf{DialoGPT} \cite{dialogpt}, \textbf{T5} \cite{T5}, and \textbf{BART} \cite{bart}. These were finetuned with RAKE keywords extracted from human-authored tongue twisters as the input and the tongue twister text from \textbf{TwistList} as the target. This was in order to represent our baselines training on in-domain data. At inference time, the prompt ``\texttt{Generate tongue twisters about the keyword(s): X}'' is used, where X refers to the input consisting of one or more RAKE keywords extracted from tongue twisters. The full training details are given in Appendix 
\ref{apx:Training-Details}. We also conducted experiments on all aforementioned baselines without finetuning (i.e., a zero-shot setting), and the results were very poor. Therefore, we did not include these results in the paper. 
\paragraph{Training-Free Baseline}
We additionally provide a TwisterMister baseline that does not require any training. We utilise OpenAI's \textbf{ChatGPT}\footnote{\url{https://chat.openai.com/chat}} with the same prompt as a zero-shot setting for generation.\footnote{No direct comparison is made to PANCETTA \cite{pancetta} as no code has been publicly released at the time of writing, and essential implementation details are absent from the paper.} Each request to ChatGPT was submitted as part of a separate session, to avoid the effects of extended dialogue influencing outputs. ChatGPT has been utilised in order to set a practical upper-bound of what may be expected from models without explicit phonetic knowledge, owing to its wealth of training data and 175B parameter architecture.\footnote{ChatGPT based on GPT-3.5, rather than GPT-4.} It is assumed that ChatGPT's training data contains tongue twisters, and therefore it is able to abstract away the general patterns of such language in order to provide novel examples (though most likely based on graphemes rather than phonemes).

\section{Experiments}

\paragraph{Automatic Evaluation.}
We present the results of automatic evaluation on generated outputs and golden examples in \autoref{tab:auto-evaluation} for the following metrics: \textbf{Perplexity} (\textbf{PPL}), \textbf{BLEU} (\textbf{B-1/B-2}) \cite{bleu}, \textbf{ROUGE} (\textbf{R-1/R-2/R-L}) \cite{rouge}, and \textbf{BERTScore} Precision, Recall, and F-Measure \cite{bert-score} (\textbf{BS-P/BS-R/BS-F}). PPL, BLEU and ROUGE are standard metrics in language generation to assess quality, whilst BERTScore assesses semantic similarity to a gold reference. 
Additionally, we propose two new metrics, Phonetic Overlap (\textbf{PO})  and Initial Phonetic Overlap (\textbf{Init-PO}). \textbf{PO} refers to the average overlap of all phonemes across tokens (\#~unique phonemes / \#~total phonemes), whereas \textbf{Init-PO} is the ratio of unique word-initial phonemes to the number of words (\#~unique word-initial phonemes / \#~words). 

These phonetic metrics reward longer outputs. We argue that, all things equal, a longer tongue twister is better than a shorter one as it provides more entertainment and more opportunities for mispronunciation. Perfect scores on PO and Init-PO can be achieved by repetition of a single word. Whilst this does not lead to high quality outputs, these metrics are intended exclusively to be indicators of the phonetics, rather than an overall guide to quality. In both cases, higher levels of overlap results in lower (``better'') scores, and the highest (``worst'') achievable score is 1.

The results in Table~\ref{tab:auto-evaluation} show rather clear scaling, with the performance ranking on most metrics (except Perplexity and phoneme overlap) being identical. On the models explicitly finetuned for this task, GPT-2 is shown to be the worst, whilst BART performs the best. We hypothesise that GPT-2's poor performance is in part due to its simple causal language modelling objective alongside its decoder-only architecture (which is also in DialoGPT). Furthermore, whilst T5 performed well on the automatic metrics, manual inspection revealed that T5 often misinterpreted the task from the prompt, choosing to select its own keywords from the entire prompt, rather than using only the provided keyword list. On the other hand, the training-free zero-shot model, ChatGPT, was shown to perform best on all metrics. This is to be expected as ChatGPT has over 50x more parameters than any other tested PLM, with various pre-training objectives and reinforcement learning, leading to performant zero-shot capabilities. This further demonstrates that PLMs struggle to learn phonetic patterns implicitly from text, especially in English, which has high levels of irregular orthography. Furthermore, with limited data, PLMs struggle to learn the unusual probability distributions underlying tongue twisters, where word choices are intentionally ``twisted'', obscure, and anti-euphonious. Additionally, due to the wealth of training data seen by ChatGPT, it is likely that many examples have been seen during training.

\paragraph{Human Evaluation.}
Due to tongue twisters being a creative domain where articulation abilities are tested, we also perform human evaluation. 3 evaluators were asked to rate 100 outputs from the best performing standard baseline (BART), in addition to ChatGPT outputs and gold examples from \textbf{TwistList} on the following criteria: \textbf{Relevance} (how relevant the tongue twister is given the keyword inputs), \textbf{Fluency} (how grammatically valid the output is), \textbf{Difficulty of Articulation} (how difficult a tongue twister is to say), \textbf{Cohesion} (how much sense the output makes), and \textbf{Entertainment Value} (how entertaining the output is, considering sounds and semantics). All ratings were on a 5-point Likert scale. Evaluator demographics and training materials are in Appendix \ref{apx:Human Participants}. 

The mean scores of human evaluation (Table~\ref{tab:human_eval_chatgpt}) fall in line with expectations, with \textit{golden} examples performing best on all metrics, and ChatGPT placing second on all but Difficulty of Articulation.\footnote{We exclude relevance for the golden examples as these were collected from the web, not elicited with keyword prompts.} BART is able to produce outputs that are deemed to be the second most difficult to articulate, which we infer may be the result of slight morphological variants of input keywords being used repeatedly, making distinguishing between them during articulation quite challenging (whilst not being able to exploit deeper phonetic relations). The moderate score on Fluency (3.028) suggests instances of poor grammar may also hinder articulation abilities when expected grammatical structures are not found, leading to an interaction between grammatical validity  and articulatory difficulty. Additionally, ChatGPT scoring the lowest for articulatory difficulty may be due to occasionally misunderstanding the requirements of a tongue twister, sometimes producing rhymes or standard prose (see Appendix \ref{Further Comments}). However, ChatGPT scores well for Relevance and Fluency, highlighting its capability in producing high-quality coherent language. Perhaps most interestingly, none of the BART score averages on any human evaluation criteria fall below 3 (``neither agree nor disagree''). This performance is therefore quite good for a model finetuned on only 2128 examples, with no additional phonetic knowledge.

% ----------- tab: human evaluation for chatgpt-----------
\begin{table}[tb]
\centering
\resizebox{0.95\linewidth}{!}{
\begin{tabular}{r|lcc}
\toprule[1pt]
\multirow{2}{*}{\textbf{Score (1 to 5)}} & \multicolumn{3}{c}{\textbf{Human Evaluation}}  \\
\cline{2-4} 
& \textbf{BART} & \textbf{ChatGPT} & \textbf{Golden}   \\
\midrule
\textbf{Relevance} & 4.667$^{*}$ & 4.971$^{\dagger}$ & N/A \\ %\textbf{4.980} 
\textbf{Difficulty of Articulation} & \underline{4.143}$^{*}$ & 4.102$^{*}$ & \textbf{4.291}$^{*}$ \\
%Formerly called Adequacy
\textbf{Fluency} & 3.028$^{**}$ & \underline{4.915}$^{**}$ & \textbf{4.938}$^{**}$ \\
\textbf{Coherence} & 3.217$^{*}$ & \underline{4.798}$^{*}$ & \textbf{4.909}$^{*}$ \\
\textbf{Entertainment Value} & 3.269$^{*}$ & \underline{4.070}$^{*}$ & \textbf{4.254}$^{*}$ \\
\bottomrule[1pt]
\end{tabular}
}
\caption{Results of Human Evaluation. The best scores are in \textbf{bold}, and the second best are \underline{underlined}. We calculate Fleiss' Kappa for each metric, and mark the agreement fair$^*$, moderate$^{**}$ and substantial$^{\dagger}$.}
\label{tab:human_eval_chatgpt}
\end{table}
% ----------- end of tab-----------

% ----------- tab: simple case study -----------
% VERSION WITH INPUT-OUTPUT PAIRS
\begin{table}[ht]
\centering
\resizebox{\linewidth}{!}{
\begin{tabular}{p{0.2\linewidth}|p{0.80\linewidth}}
\toprule
\textbf{Input} & assistant assist \\
% \textbf{Phonemes:} & [AH0 S IH1 S T AH0 N T] [AH0 S IH1 S T] \\
\midrule
\textbf{GPT-2} & assistant assist assistant assist assistant \\
\midrule[1pt]
\textbf{DialogGPT} & assistant assistant assistant assistant assistant assistant assistant assistant  \\
\midrule
\textbf{T5} & assistant assist assistant \\
\midrule
\textbf{BART} & A assistant assist is an assistant assist, assistants assist to assist assistants.\\
\midrule
\textbf{ChatGPT} & Assistant ants assist ants in carrying leaves to the ant hill. \\
\midrule
\textbf{Golden} & If I assist a sister-assistant, will the sister's sister-assistant assist me? \\
\bottomrule
\end{tabular}
}
\caption{Example outputs for the input "assistant assist". "Golden" refers to the human-authored tongue twisters.}
\label{tab:case_study}
\end{table}

\section{Case Study}

Within the example in \autoref{tab:case_study}, GPT-2 resorts to simply repeating the input, successfully achieving phonetic overlap, but failing to be grammatically valid or particularly sophisticated. This pattern is also demonstrated by DialoGPT and T5. Conversely, BART is able to introduce tokens unseen in the input to create an almost grammatically valid output (the primary mistake being indefinite article agreement, where in the first instance ``an'' would have been correct, rather than ``a''). BART's output is also semantically and logically coherent, with ``A assistant assist is an assistant assist'' being valid (yet redundant), and ``assistants assist to assist assistants'' also being comprehensible. This example demonstrates why evaluators with high English proficiency and language/linguistics education were selected, as the same word may have different parts of speech, creating outputs that seem grammatically invalid, but do actually follow the rules of English.\footnote{\url{https://en.wikipedia.org/wiki/Buffalo_buffalo_Buffalo_buffalo_buffalo_buffalo_Buffalo_buffalo}} Further investigation is needed to ascertain whether the models are intentionally exploiting this lexical ambiguity, or if human evaluators are demonstrating apophenia, where patterns are found in what is effectively noise \cite{Brugger_2001}. Finally, ChatGPT utilises morphology to exploit the similarity of the plural noun ``assistants'' and the phrase ``assist ants'', and provides a continuation that is in line with the expected behaviour of ants. In comparison to the golden example, ChatGPT's output may be considered more interesting topic-wise, at the expense of not being as phonetically complex (``carrying leaves to the ant hill'' contributes heavily to semantics, whilst not being recognisable as part of a tongue twister). For further analysis, please see Appendix \ref{Further Comments}.

% % ----------- end of tab-----------
% =============================== Section 6 ==================================
\section{Conclusion}
We present work on the topic of tongue twister generation, a form of phonetically-constrained language generation that aims to maximise phonetic overlap, whilst conveying meaningful semantics. We motivate the potential application domains for such generated language, and provide a large annotated dataset of tongue twisters, \textbf{TwistList}, to encourage further work. Finally, we present a series of benchmark models alongside automatic/human evaluation to assess generation quality.

\section*{Limitations}
Whilst the system presented within this paper is capable of allowing human-in-the-loop contributions (via selecting the input keywords on which to condition the output), it is not able to produce tongue-twisters that take advantage of particular features of speech sounds such as place and manner of articulation, in order to create more advanced outputs that exploit phonetic relatedness (rather than exact matches). The same can be said of our proposed metrics, PO and Init-PO, which do not account for phonetic similarity across sounds that share manner/place of articulation (e.g. "\textbf{sh}e \textbf{s}ells \textbf{s}ea \textbf{sh}ells"). Additionally, whilst commonly known tongue twisters may follow a particular format (e.g. rhyme schemes), such schemes and templates have not been enforced here. We also do not demonstrate the capabilities of these systems if they were trained on phonetic transcriptions explicitly, as we only aim to assess their performance when training on graphemes in standard orthography.  

\section*{Ethics Statement}
All use of human participants in this study has been approved by the Ethics Board of the primary author's institution, including the disclosure of demographic information. Regarding the generation of tongue twisters, language generation is a necessarily creative domain that has the ability to reproduce content that some individuals may find offensive. Care was taken to check outputs in the human evaluation set for any such materials, and if they had been produced, they would have been removed from the evaluation set. Additionally, no egregiously offensive material has been provided in the TwistList dataset. However, the distinction between offensive and humorous content is a highly complex topic, and therefore some examples within the dataset may not be suitable for all individuals (e.g. suggestive content and swearing, such as "I'm not the pheasant plucker, I'm the pheasant plucker's son", and the clear relation to common expletives).

\section*{Acknowledgements}
Tyler Loakman is supported by the Centre for Doctoral Training in Speech and Language Technologies (SLT) and their Applications funded by UK Research and Innovation [grant number EP/S023062/1]. Chen Tang is supported by the China Scholarship Council (CSC) for his doctoral study (File No.202006120039). 
% % =============================== Reference ==================================
\bibliographystyle{acl_natbib}
\bibliography{custom}

% \section*{Acknowledgements}

% =============================== Appendices ==================================
\appendix
\section{Appendices}
\label{sec:appendix}
\subsection{Dataset Quality Control}
\label{apx:QC}
\begin{figure*}[t]
\centering
\includegraphics[width=0.95\linewidth]{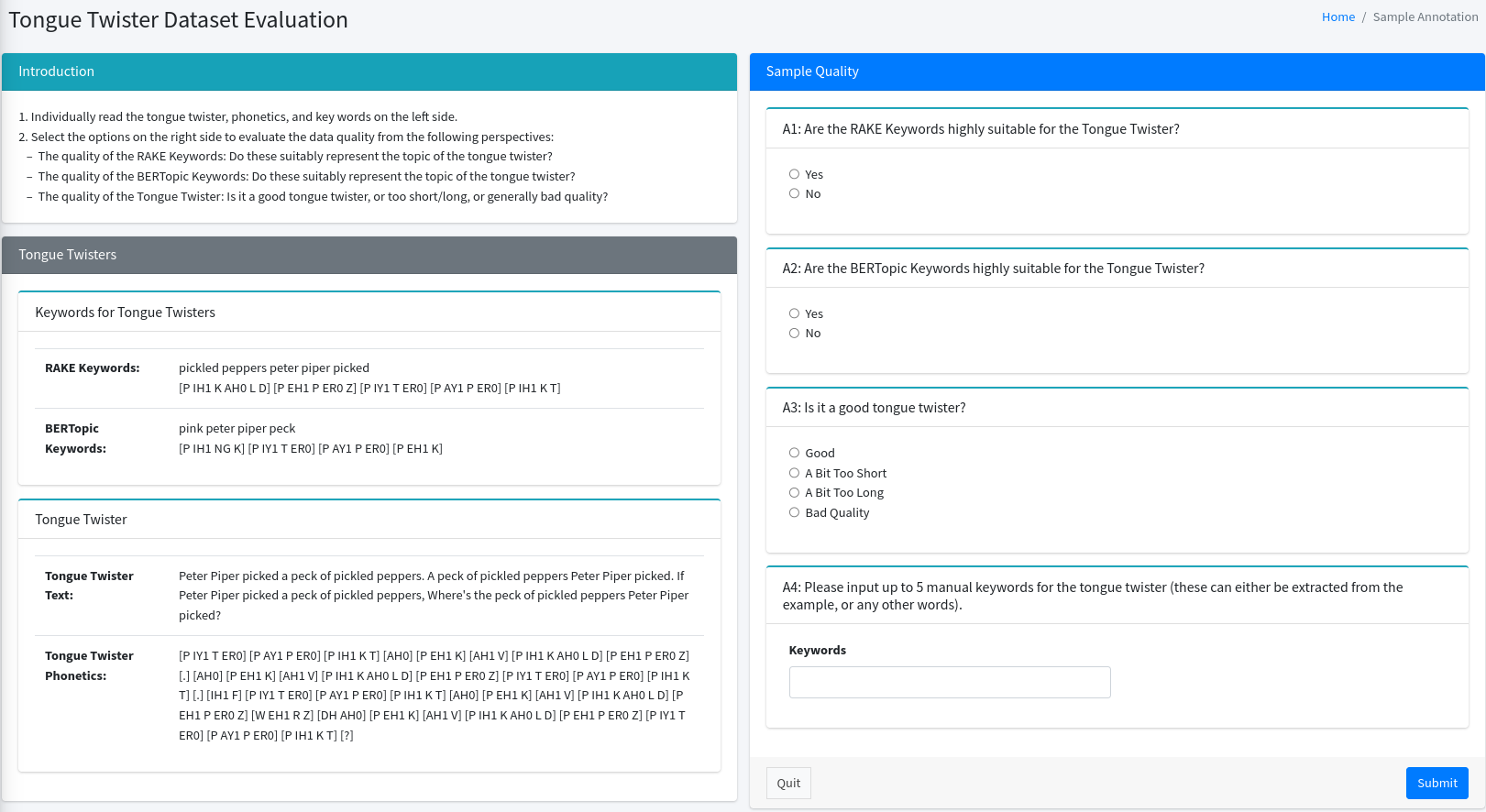}
% \includesvg[width=1.0\columnwidth]{figs/intro.svg}
\caption{TwistList Quality Control Annotation Platform} 
\label{fig:QC}
\end{figure*}

An annotation platform was developed as shown in (\autoref{fig:QC}).

% ----------- fig:overview -----------
% ----------- end of fig -----------

\subsection{Human Participants}
\label{apx:Human Participants}
Due to tongue twisters being highly reliant on articulation abilities, the demographics of the human participants used within this work are highly important. Additionally, tongue twisters are also a form of humour and entertainment, where individual perceptions of what may or may not be considered humorous or entertaining differ according to numerous factors. In an effort to remain as transparent as possible, and follow best practices for human evaluation, relevant demographic information of participants are outlined below (with the necessary requisite permission and ethical approval).

\paragraph{Dataset Evaluation}

All evaluators involved in the quality control process of the \textbf{TwistList} dataset are native speakers of English, and either have or are working towards University level qualifications. Additionally, 2 of the 3 evaluators have extensive education in linguistics or modern languages. No monetary incentive was provided.

\paragraph{Generation Evaluation}

All evaluators involved in the evaluation of the quality of generated tongue twisters are native speakers of English, and either hold or are working towards University level qualifications in Linguistics, Modern Languages or NLP. Additionally, all evaluators cited the United Kingdom as their country of socialisation, and no participants reported language processing difficulties that could affect results. No monetary incentive was provided.

\begin{figure*}[t]
\centering
\includegraphics[width=0.95\linewidth]{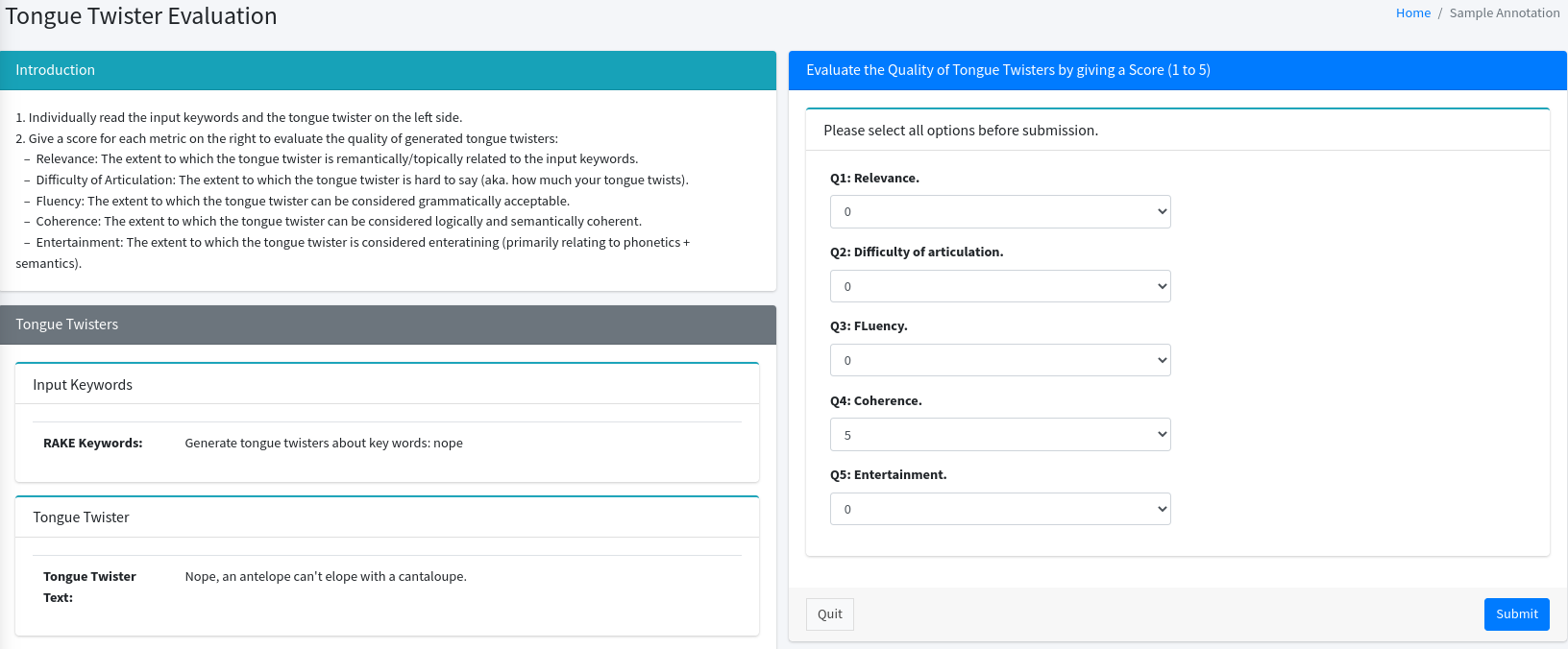}
% \includesvg[width=1.0\columnwidth]{figs/intro.svg}
\caption{Human Evaluation Platform for Generated Outputs} 
\label{fig:eval_platform}
\end{figure*}

\paragraph{Materials Provided to Human Participants}
Additionally, all evaluators for both the dataset and generation outputs were presented with calibration examples to demonstrate the sort of outputs that would be presented, and the logic behind particular scores, in order to minimise individual interpretations of the scoring criteria. All evaluation was performed on a custom made online annotation platform (\autoref{fig:eval_platform}).

% ----------- fig:overview -----------
% ----------- end of fig -----------

% \subsection{Phonetic Metrics}
% \label{apx:Phonetic Metrics}
% Within this paper we present \textit{Init-PO} and \textit{PO}. Init-PO is the word-initial phoneme overlap across tokens, and is equal to the number of unique word-initial phonemes divided by the total number of word-initial phonemes (which is the same as the number of words). On the other hand, PO is the overall phoneme overlap, regardless of word-position, and is equal to the total number of unique phonemes within an output divided by the length of the output in phonemes. 

% Importantly, these metrics do not account for phonetic similarity across different sounds that share manner/place of articulation (e.g. "\textbf{sh}e \textbf{s}ells \textbf{s}ea \textbf{sh}ells"). Additionally, these metrics reward longer outputs. We argue that, all other things equal, a longer tongue twister is better than a shorter one. Additionally, perfect scores on these metrics can be achieved by simple repetition of a single word. Whilst this does not lead to a high quality tongue twister, these metrics are intended exclusively to be indicators of the phonetics of a tongue twister, rather than an overall guide to quality.

% In both cases (Init-PO and PO), higher levels of overlap results in lower ("better") scores, and the highest ("worst") achievable score is 1.

\subsection{Training Details}
\label{apx:Training-Details}
All pre-trained models used (naturally excluding ChatGPT) are based on publicly available checkpoints from Hugging Face.\footnote{\url{https://huggingface.co/models}} Models are trained for up to $5$ epochs on a Tesla A5000 machine  with the best checkpoints selected based on the validation loss. The batch size is set to $32$, and the learning rate is $8e^{-5}$, with the Adam optimiser selected for training. To help the loss curve converge on our small few-shot dataset, we limit the generation length to $100$ (covering all test tongue twisters). Meanwhile, the source length is limited to $150$. The training and testing steps are set up with the implementation of the PyTorch Lightning\footnote{\url{https://www.pytorchlightning.ai/}} framework to guarantee the reliability of the experiment. All language models are fairly trained and tested with the same steps.

% \subsection{Additional Automatic Metrics}
% \autoref{auto-evaluation-appx} presents further automatic evaluation on generated tongue twisters, specifically higher-order BLEU (\textbf{B-3/B-4}).
% % ----------- tab: metrics -----------
% \begin{table}[ht]
% \centering \small
% \resizebox{0.98\linewidth}{!}{
% \begin{tabular}{l|ccccc}
% \toprule
% \textbf{Model}  & \textbf{B-3$\uparrow$} & \textbf{B-4$\uparrow$} & \textbf{PO-S} & \textbf{PO} & \textbf{Avg. Len} \\
% \midrule
% \textbf{GPT-2} & 0.002 & 0.000 & 0.142 & 1.230 & 101.1 \\
% \textbf{DialoGPT} & 0.018 & 0.013 & 0.267 & 1.140 & 77.9\\
% \textbf{T5} & 0.024 & 0.014 & 0.681 & 0.820 & 11.2 \\
% \textbf{BART} & 0.037 & 0.025 & 0.721 & 0.730 & 96.1 \\ 
% \midrule
% \textbf{ChatGPT} & \textbf{0.102} & \textbf{0.077} & 0.679 & 0.677 & 10.18 \\ 
% \bottomrule
% \end{tabular}
% }
% \caption{\label{auto-evaluation-appx}
% Additional Automatic Evaluation. The PO score of the golden examples is 0.760. 
% }
% \end{table}
% % ----------- end of tab-----------

\subsection{Further Qualitative Comments}
\label{Further Comments}
Whilst the pattern of extreme word repetition is seen in many of the finetuned models (often with the exception of BART, which is demonstrated to be capable of producing slightly more sophisticated outputs), overall assessment of the tongue twisters produced at inference time reveals interesting patterns, particularly in regard to ChatGPT outputs. Firstly, the limits of ChatGPT are made apparent in a few examples such as the input "silver shiny ship sank" generating "How much wood would a woodchuck chuck if a woodchuck could chuck silver shiny ships?", a clear derivation of a famous woodchuck related tongue twister that it is rather safe to assume appears multiple times in ChatGPTs training material. Additionally, comments from evaluators also reveal that ChatGPT's output is often considered more of a rhyme or general literary text, rather than specifically a tongue twister. However, examples such as these are also found in the human-authored golden examples, demonstrating that there is no steadfast consistent opinion as to what constitutes a (good) tongue twister. Likewise, some examples may contain large amounts of sound repetition, but not in a way that necessarily presents articulatory difficulty.

\subsection{Future Works}
In this paper, we mainly analyse the performance of large-scale pretrained language models (PLMs) on Tongue Twister Generation, and propose a corresponding dataset for further investigation. In further works, we aim to propose novel models which can better leverage phonetic symbols. There are numerous existing works \cite{huang-etal-2022-improving, tang-etal-2022-etrica, tang2022terminology,zhanga2023cadge} that provide approaches for injecting such knowledge into PLMs. However, the phonetic features differ from these text-format knowledge items, as phonemes are hard to encode with input text tokens when feeding into PLM encoders. Another promising approach is to explicitly model the phonetic features into text sequences \cite{tang2022ngep}, though there is no observed method for transforming phonetic notation. We intend to perform further research based on these existing approaches.

\end{document}